\documentclass[preprint]{format/vgtc/vgtc}                 

\ifpdf
  \pdfoutput=1\relax                   
  \usepackage{graphicx}                
  \DeclareGraphicsExtensions{.pdf,.png,.jpg,.jpeg} 
\else
  \usepackage{graphicx}                
  \DeclareGraphicsExtensions{.eps}     
\fi%

\graphicspath{{figures/}{pictures/}{images/}{./}} 

\usepackage{microtype}                 
\PassOptionsToPackage{warn}{textcomp}  
\usepackage{textcomp}                  
\usepackage{mathptmx}                  
\usepackage{times}                     
\usepackage{cite}                      
\usepackage{tabu}                      
\usepackage{booktabs}                  
\usepackage{kotex}
\usepackage{subfig}
\usepackage{amsmath} 
\usepackage{algorithm}
\usepackage{algorithmicx,algpseudocode}
\usepackage{gensymb}
\usepackage{textgreek}

\vgtccategory{Research}

\vgtcinsertpkg

\title{eCAR: edge-assisted Collaborative Augmented Reality Framework}

\author{Jinwoo Jeon\thanks{e-mail: zkrkwlek@kaist.ac.kr} %
\and Woontack Woo\thanks{e-mail:wwoo@kaist.ac.kr} %
}
\affiliation{\scriptsize KAIST GSCT UVR Lab. \\ KAIST GSCT UVR Lab.}

\teaser{
 \centering
  \includegraphics[width=\linewidth]{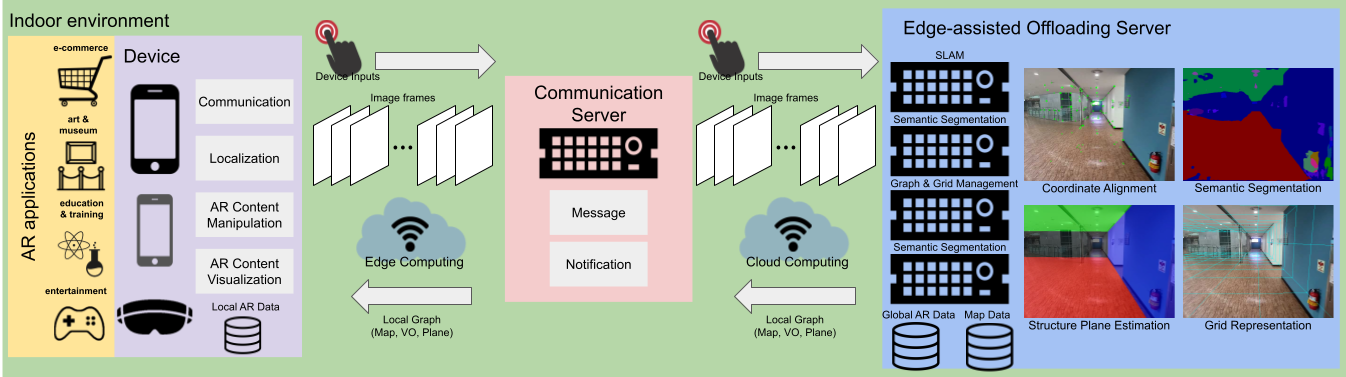}
 \caption{Concept of our proposed system.}
 \label{fig:teaser}
}

\abstract{
We propose a novel edge-assisted multi-user collaborative augmented reality framework in a large indoor environment.
In Collaborative Augmented Reality, data communication that synchronizes virtual objects has large network traffic and high network latency.
Due to drift, CAR applications without continuous data communication for coordinate system alignment have virtual object inconsistency.
In addition, synchronization messages for online virtual object updates have high latency as the number of collaborative devices increases.
To solve this problem, we implement the CAR framework, called eCAR, which utilizes edge computing to continuously match the device's coordinate system with less network traffic.
Furthermore, we extend the co-visibility graph of the edge server to maintain virtual object spatial-temporal consistency in neighboring devices by synchronizing a local graph.
We evaluate the system quantitatively and qualitatively in the public dataset and a physical indoor environment.
eCAR communicates data for coordinate system alignment between the edge server and devices with less network traffic and latency.
In addition, collaborative augmented reality synchronization algorithms quickly and accurately host and resolve virtual objects.
The proposed system continuously aligns coordinate systems to multiple devices in a large indoor environment and shares augmented reality content.
Through our system, users interact with virtual objects and share augmented reality experiences with neighboring users.
}

\CCScatlist{
  \CCScatTwelve{Augmented Reality}{Sharing Contents}{Multi-user AR}
  \CCScatTwelve{Collaborative Augmented Reality}{Edge-assisted AR}
} 

\begin{document}
\def\cogr{co-visibility graph}
\def\commumap{eCARAA}
\def\baselmap{eCAR-LocalMap}
\def\mpmethodone{map point decoupling}
\def\mpmethodtwo{map point selection strategy}

\firstsection{Introduction}
\maketitle
With the recent development of technology, multi-user AR applications have bounced in interest among smart device users.
Multi-user AR applications visualize virtual objects that one user registers on the display of other adjacent users.\cite{spar, sharear, edgearx5}
Google, Apple, and Microsoft offer collaborative augmented reality platforms\cite{ARKit, SpatialAnchor, cloudanchor, MetaAnchor}.
In collaborative augmented reality, to visualize a virtual object in the same location, devices synchronize map information and virtual object information to the network.

We define the process of overlaying virtual objects on the screen of the device in Collaborative AR in five steps as follows.
The first step is initialization communication.
Initialization communication communicates visual clue data that matches coordinates between two devices.
In the communication process, the device generates approximately 9 to 20 MB of traffic during coordinate system matching, and the latency is about a few seconds\cite{spar}.
The second step is real-time device location tracking.
In general, the location tracking step is a sensor-based approach using the Inertial Measure Unit(IMU) or a vision-based approach using feature points in an image.
Vision-based tracking is vulnerable to rapid motion changes and a texture-less environment.
In addition, mapping optimization is unsuitable for mobile devices with limited performance due to large computations.
Maps without mapping optimization may cause drift.
Sensor-based tracking using a gyro sensor and an accelerometer sensor tracks the device's pose relatively accurately with a small amount of calculation on high-movement devices.
However, due to the properties of the sensor, tracking accuracy is reduced if the device does not move.
In addition, the gyro sensor accumulates errors over time due to its integral characteristics.
The fourth step is to understand the physical space.
To this end, mobile augmented reality recognizes real objects or inferences semantic information from sequences.
Collaborative augmented reality recognizes physical objects and overlays virtual annotations on them.
However, physical spatial understanding requires much computation, so it offloads the detection module to the edge server.\cite{edgeObjectAR,sear}
The latency that it takes from the device to transmit the image and receive the result is quite long, requiring a data conversion process.\cite{edgeSemanticSLAM1}
The final step is visualization and rendering.
Mobile augmented reality visualizes various information on a screen.
The information augmented in the physical world is calculated to be visible at the same location according to the device location change.
According to Valentin et al., visualization of 3D information uses mobile GPUs, so computational tasks using GPUs are unsuitable for optimization in mobile\cite{depthfrommotion}.

In mobile, collaborative AR considers communication data traffic and latency by edge or cloud computing.
Matching coordinate systems frequently is difficult because of huge network traffic.
However, the absence of a coordinate system alignment trial causes spatial drifts or inconsistencies over time.
Due to large network traffic to align coordinate systems, CAR can only match narrow areas around virtual objects. 
In addition,  a collaborative AR framework requires communication between devices to maintain the same virtual object state.
Ruth et al.\cite{car_quad} argued that as the number of users and objects increases, the number of messages for interaction increases quadratically.
Collaborative augmented reality frameworks that update online virtual objects require more interaction messages.
Such an increase in messages causes bottlenecks and reduces augmented reality consistency.\cite{edge_bottleneck}

Latency and spatial consistency\cite{carkeyfactor1,carkeyfactor2,carkeyfactor3} are the main factors of sharing experiences in collaborative augmented reality.
The high latency by data communication of existing collaborative AR platforms \cite{ARKit, SpatialAnchor, cloudanchor} is a direct factor that reduces spatial consistency\cite{spar}.
Edge-assisted SLAM\cite{edgeSLAM1,edgeSemanticSLAM1} is an appropriate method for matching coordinate systems between devices for virtual object visualization.
These studies communicate local maps between an edge server and devices.
However, like existing collaborative augmented reality platforms, their communication generates much network traffic.
FreeAR\cite{freear} and eAR\cite{ear} suggest an energy-efficient mobile AR framework using an edge-computing mechanism.
Edge Robotics\cite{edgerobotics} proposed an edge computing-based collaborative SLAM using LiDAR for multi-robot applications with low latency.
We propose an algorithm that continuously matches coordinate systems for collaborative augmented reality experiences that maintain spatial consistency.
Decentralized collaborative AR frameworks communicate data without additional resources\cite{freear, spar}.
These approaches communicate data between adjacent devices directly\cite{uwb,bluetooth,wifi_direct}.
However, as the number of collaborative devices increases, decentralized communication devices are burdensome to communicate with each other.
Also, direct communication in decentralized CAR may be possible to communicate data from circulating indefinitely.

To remove these limitations of CAR, this research utilizes edge-computing SLAM\cite{edgeSLAM1, edgeSemanticSLAM1}.
Regardless of the advantage of edge-computing SLAM, working collaborative AR with communication is a significant problem.
First, the SLAM map communication for state consistency among the edge server and devices causes huge network traffic.
In addition, frequent sharing of the SLAM map status on multiple devices causes network traffic drastically.
Second, unsuited SLAM for virtual object synchronization
All virtual object interaction synchronization, such as registration and modification in the SLAM map, affects the network condition.

To solve these problems, this research proposes eCAR, an edge-computing system built upon the graph-grid-data structure to enable collaborative augmented reality between multiple co-located devices in low-traffic data communication. 
This graph-grid-data structure can communicate map data and virtual object data at once.
Our motivations are as follows:
\begin{itemize}
    \item Integrated coordinate system management in the edge server provides the same coordinate system to the device without drift.
    \item Communication using a co-visibility graph shares a local graph to a device with less computation.
\end{itemize}
Using them, eCAR considers an edge-assisted centralized communication system, where each device individually communicates data that can maintain state with the edge server.
The edge server guarantees achieving SLAM with received data from devices, aligning the coordinate system of devices, and synchronizing the local graph with devices for AR consistency. 
Furthermore, the planar-based grid representation of an indoor environment facilitates virtual object synchronization without physical limitations such as distance and angle.
Figure \ref{fig:teaser} describes our proposed CAR framework, called eCAR.
When eCAR receives an image from a device, this framework aligns the device's position and finds proper AR contents connected with grid cells.
Then, eCAR sends a local graph, including map data and AR data, to the device.

We implement eCAR, server, and client on real devices and simulations.
Experimental results show that our proposed system can outperform the state-of-the-art solution by up to 370\% traffic and 62\% latency reduction on multiple devices when the number of connected devices is up to 20.
Our CAR drawing application, used for proof-of-concept in an indoor environment, verifies its efficiency and feasibility and shows the advantages of our edge-assisted collaborative AR framework.

In this study, our contributions are as follows:
\begin{itemize}
\item We propose eCAR, a collaborative Augmented Reality framework based on the graph-grid-data structure to enable low-latency graph synchronization. 
To connect data with grid share virtual objects selectively without distance and angle limitation in a large indoor environment.
\item We design a communication mechanism that shares SLAM map points and virtual objects for AR consistency among multiple devices.
\item We develop novel graph-based SLAM map communication with low input and output traffic for coordinate alignment in the edge server.
Additionally, we suggest parameters for collaborative AR experiences that communicate with low latency, maintain localization accuracy, and minimize network traffic.
\item We implement and evaluate eCAR in simulation and real devices for scalability. 
This simulation evaluation proves the effectiveness of the eCAR framework and its communication mechanism with up to 20 clients.
Besides, real device evaluation shows AR interaction among four devices simultaneously.
\end{itemize}  
\section{Related work}
Due to hardware limitations, mobile augmented reality matches coordinate systems with edge or cloud computing.
Edge-assisted SLAM\cite{edgeSLAM1, edgeSemanticSLAM1} 
studied SLAM module decoupling.
These studies transfer keyframes from the device to the server and local maps from the server to the device.
The device estimates the location of the device using a local map transmitted from the server.
Local map transmission has large network traffic.
To reduce this, edgeSLAM\cite{edgeSLAM1, edgeSLAM2} have studied the transmission period and method of local maps.
edgeSLAM\cite{edgeSemanticSLAM1} sends only changed map points.
C2TAM\cite{c2tam} proposed a method of offloading modules with cloud-computing. 
C2TAM communicates an entire global map.
Schmuck et al.\cite{multi_uav,ccm_slam} sends a local map to the device whenever the server performs bundle adjustment. 
They do not align coordinate systems and still maintain relative coordinate systems.
ORB-SLAM3\cite{orb_slam3} proposed an improved place recognition method to match new maps with poor visual information over long periods of time on a single device.
SPAR\cite{spar} aligns a coordinate system around a virtual object.
This study proposed a method to measure spatial inconsistency.
SLAM-share\cite{slam_share} is localized on a server to match a coordinate system. 
However, it transmits only a pose of a device, not a local map. 
A device estimates pose only with an IMU sensor.
MARVEL\cite{marvel} returns the pose of a device when it sends an image to a server.
This study communicates already created maps only once initially without optimization.
SynchronizeAR\cite{synchronizar} estimated the location between the two devices without map matching by attaching additional hardware communicating with Ultra Wide Band (UWB).
AVR\cite{avr} studied a communication method for matching a coordinate system with 3D point clouds between two autonomous vehicles while reducing latency.
They communicate motion vectors with point clouds compressed with lossless compression algorithms.

Edge-assisted collaborative AR studies recognize objects or images to align coordinate systems on mobile\cite{edgear1,edgear2,edgear3,edgear4}.
Overlay\cite{overlay} transmits the original image to the server, detects feature points with SURF from the server, and builds a DB for image retrievals. 
They operate on images of poor quality due to the resolver's blur or hand motion.
The mobile visualizes an annotated object with sensor information.
VisualPrint\cite{visualprint} defines the most distinctive visual data as visual fingerprints and has been offloaded to the server.
They detect and compress keypoints on the device.
Thereafter, compressed keypoint information is transmitted to the server to reduce network traffic.  
JAGUAR\cite{jaguar} proposed an approach that combines existing AR systems with image retrievals utilizing edges.
They recognize objects by edges.
On the other hand, they estimate device locations with augmented reality frameworks such as ARCore or ARKit.
This study does not share maps between devices.

Image-based coordinate systems share relatively few features around the image compared to an environment, which is advantageous for continuous coordinate system alignment in certain areas.
However, this approach is unsuitable for coordinating system alignment in large indoor environments.

Collaborative AR has been studied data communication for interaction between devices \cite{car1,car2,car3,car4,car5,car6}.
Collaborative AR communicates between adjacent devices to align coordinate systems that visualize the same virtual object.
In addition, CAR communicates data that synchronizes interaction messages.
CAR\cite{cars} reduced latency by exchanging object recognition results with device-to-device (d2d) communication, taking into account human properties in indoor environments such as museums.
ShareAR\cite{sharear} defined problems for sharing AR experiences and presented Quality of Augmentation to solve these problems.
EdgeAR X5\cite{edgearx5} defines a communication plan for synchronizing multi-user interactions.
They collaborated with nearby mobile devices through d2d communication to reduce initialization latency.
COMIC\cite{comic} proposed an infrastructure with a variety of integrated features to provide immersive experiences to physically remote multi-users.
However, they do not address the construction of an entire integrated system, taking into account optimized combinations and collisions between individual solutions.

In augmented reality, a spatial anchor serves as a coordinate system for fixing virtual objects in the physical world\cite{ARKit,ARCore,MetaAnchor,SpatialAnchor}.
In order to visualize virtual objects in collaborative augmented reality, an understanding of the real world is essential.
Based on this, the anchor aligns the two coordinate systems.
After alignment, each device visualizes the same virtual object.
However, the physical limitations of the spatial anchor that can share virtual objects are 5 meters.
Thus, a coordinate system in CAR that is coupled with an anchor synchronizes virtual objects in a limited area.
To solve this problem, the Spatial Anchor added a new anchor and connected it with the existing previous anchor\cite{AzureWayFinding}.
However, this method requires a user effort whenever the anchor is connected.
We propose a novel algorithm based on edge computing that continuously aligns coordinate systems with less network traffic for collaborative augmented reality.
\section{Background}
In this section, we briefly describe the general in-direct SLAM\cite{ptam,orb_slam1,orb_slam2}.
Typically, SLAM has three modules: tracking, local mapping, and loop-closing.
The tracking module estimates the position of the camera at the front end.
The local mapping and loop-closing modules optimize local or global maps at the back end\cite{ptam}.
The data structure of SLAM has mappoints and keyframes.
These data are connected by \cogr \cite{cg_vis_graph}.
A local map is a set of map points observed by keyframes that share observed mappoints in a frame.
The tracking module acquires more matching pairs between a local map and the current frame to increase camera pose estimation accuracy.
Local mapping optimizes using \cogr to minimize the repetition errors of keyframes and map points connected to new keyframes called bundle adjustment\cite{bundle_adjustment,g2o,ceres_solver}.
In this paper, eCAR augments \cogr properties to communicate the proper AR state in the edge server to multiple devices.

Edge-assisted SLAM offloads computational-intensive tasks to servers due to device limitations\cite{edgeSLAM1, edgeSLAM2, edgeSemanticSLAM1, edgeSemanticSLAM2}.
They analyze and decouple the task for a mobile SLAM.
They offload local mapping and loop-closing modules to the server, which are generally difficult to process in real time and require much computation.
The tracking module for localization remains on the device.
Edge SLAM communicates a local map between the device and the server to maintain map state consistency.
Basically, Edge-assisted SLAM generate mappoints in the device and optimize these mappoints in the edge server.
These approaches\cite{edgeSemanticSLAM1, edgeSLAM1} also implement the module based on an indirect SLAM\cite{orb_slam2}.
A semi-dense map of direct SLAM\cite{lsdslam,dso,svo} or RGB-D SLAM\cite{dtam,kinectfusion,bundlefusion} is not suitable for edge-assisted SLAM by huge network traffic.
\section{Overview of eCAR}
\begin{figure}
\begin{center}
\includegraphics[width=1.0\linewidth]{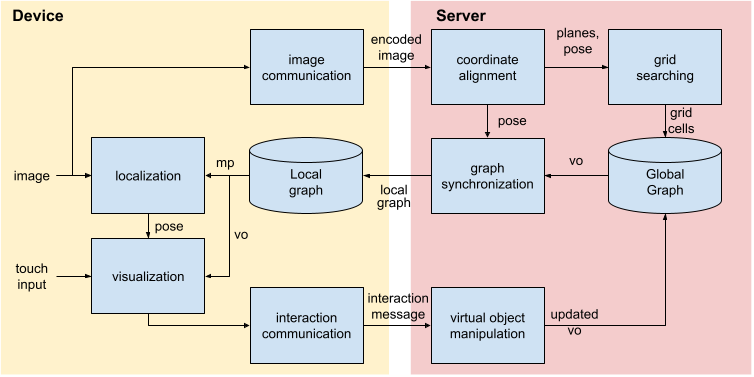}
\end{center}
\caption{System Overview.}
\label{fig:overview}
\end{figure}
This study is a system that offloads mapping modules to the edge server based on ORB-SLAM\cite{orb_slam1}.
The system uses monocular RGB images to track multiple devices in the same coordinate system on the edge server and interact with virtual objects using edge computing\cite{edgeSemanticSLAM1, edgeSLAM1}.

eCAR integrates coordinates to maintain the identical augmented reality state between multiple devices and consists of a server that communicates data and a device that tracks camera pose and interacts with virtual objects with the information provided by the edge server.
For efficient AR state sharing, eCAR modifies \cogr to connect virtual objects with maps.
In addition, eCAR manages virtual objects by adding grid cells to \cogr for AR consistency.
Figure \ref{fig:overview} shows an overview of eCAR.
The server side of eCAR has four components: the coordinate system alignment module, the grid connection module, the communication module, and the global data graph. 
Global graphs include virtual objects, plane information such as floor and wall, map points, and keyframes.
The communication module receives images to match a coordinate system and interaction messages to manipulate a virtual object from the device.
The coordinate system alignment module utilizes the tracking module of ORB-SLAM\cite{orb_slam1} to match the coordinate system of the device with the coordinate system of eCAR.
The coordinate system alignment module estimates the device's position from the coordinate system of the eCAR with images transmitted from the device.
In addition, after alignment, the communication module transmits the local graph, including mappoints and virtual objects, to the device.
The graph synchronization module temporarily connects the device to communicate data for AR consistency.
The grid connection module connects the new keyframe generated from the mapping module and grids.
The virtual object manipulation module updates the virtual object from the message and connects it to the graph.

The device side of eCAR also has four modules: the localization module, the visualization module, the communication module, and the local graph. 
The device tracks its position in the same coordinate system as the server and interacts with virtual objects in real-time. 
The device continuously communicates with the server to synchronize the local graph, which maintains AR consistency, such as map data for tracking and virtual objects for interaction. 
The communication module uploads the frame and interaction messages of the device to the server and downloads the local graph.
The localization module tracks the device's position in real-time.
The visualization module visualizes the virtual objects connected to the local graph by reflecting the device's pose.
Upon detecting a user input, the device transmits an interaction message to the edge server to manipulate the virtual object.

\begin{figure}[t]
\begin{center}
\includegraphics[width=1.0\linewidth]{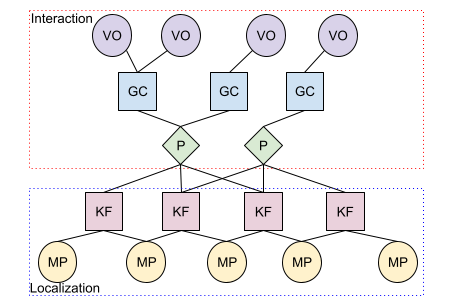}
\end{center}
\caption{Proposed Graph-Grid-Data Structure. VO is a virtual object. KF is a keyframe, MP is a map point. Observations connect KF and MP. Graph connects grid cells observed in the viewpoint of the keyframe using structure planes and the frame's pose. By user interaction input, virtual objects are connected with a grid cell in \cogr for synchronization.}
\label{fig:graph_concept}
\end{figure}
\section{Augmented Reality States Synchronization}
\subsection{Local Graph Communication}
The global graph $G^E=\{M^E,K^E,V^E,P^E,C^E\}$ includes map data $M^E$ and keyframes $K^E$ and AR content data $V^E$ and Planes $P^E$ and grid cells $C^E$.
The map data $M^E = \{m_1, m_2,...,m_{Nm}\}$ and content data  $V^E = \{v_1, v_2,...,v_{Nv}\}$ are synchronized among the edge server and devices where $m_i$ is a map point and $v_j$ is a virtual object.
Key Frames $K^E = \{k_1, k_2,..., k_{Nk}\}$ connects map points and other information indirectly using properties of the co-visibility.
Planes $P^E = \{p_1, p_2,...,p_{Np}\}$ and Grid cells $C^E = \{c_1, c_2,...,c_{Nc}\}$ facilitate augmented reality interaction.
eCAR keeps the AR states of devices almost similar to the AR states of the server.
\begin{equation}
    G^E\approx G^D
    \label{eq:appr_graph}
\end{equation}
, where $G^D$ is a graph of a device.
The mapping module in the edge server generates new map points and connects to the global graph of the server $G^E$
The virtual object management module in the edge server updates virtual objects by reflecting the user's interaction and connects to the global graph of the server $G^E$.
The edge server continuously communicates a local graph $G^L=\{M^L, V^L, P^L\}$, including map and AR data, to devices to maintain AR consistency.
Each device constructs a graph $G^D$ using the received graph $G^L$  from the edge server.
In this section, we describe the construct of a local graph $G^L$.
Periodically, the device transmits lossy compressed images to the edge server.
In this paper, devices send encoded data to every fourth frame to communicate a local graph. 
The edge server aligns the coordinates among devices using these compressed images.
The alignment process is similar to the tracking process in visual SLAM.
We apply thread pools to match images transmitted by multiple devices without bottlenecks.
The alignment process finds correspondence between the device's frame and mappoints.
Using \cogr properties\cite{orb_slam1}, the alignment process finds keyframes $K^L$. 
Lastly, eCAR finds structure plane set $P^L$ connected with local keyframes  $K^L$.
In this paper, we estimate structure planes using semantic segmentation\cite{mit_seg} and RANSAC\cite{ransac} to classify structures such as floor, wall, and ceiling from a new keyframe.
Then, eCAR connects structure planes with this keyframe.

After estimating the device's pose $T_{cw}$ and finding plane set $P^L$, eCAR searches viewing grid cells to share virtual objects.
In a grid cell, virtual objects are connected by user interaction. 
Firstly, eCAR reconstructs a 3D point $X_w$ corresponding to image $x = (u,v)$ with local structure plane set $P^L$ and transformation matrix from camera to world coordinate $Twc = [R_{wc}|t_{wc}]$  by :
\begin{equation}
l_w(u,v)=R^T K^{-1}(u,v,1)^T - R^T t
\label{eq:back_proj_lay}
\end{equation}
\begin{equation}
d(u,v) = \frac{-(d_p+O_w\cdot n_p)}{l_w(u,v) \cdot n_p}
\label{eq:dist_c2p}
\end{equation}
\begin{equation}
    X_w(u,v) = \underset{d_i}{\arg\min} O_w+d_il_w(u,v)  , if d_i < th_{dist}
\label{equ:p_recon}
\end{equation}
, where $O_w$ is the camera center position in the server coordinate, $n_p$ is the plane's normal vector, and $d_p$ is the distance from the origin to the plane.
eCAR samples uniformly distributed pixels to reduce computation for planar reconstruction. 
In this paper, we set $th_{dist}$ = 1.2 and the sample pixel window size = 5.
equ (\ref{eq:grid_convert}) project 3D point to the grid cell.
\begin{equation}
c(X,Y,Z) =  (\lfloor X/cellsize \rfloor,\lfloor Y/cellsize \rfloor,\lfloor Z/cellsize \rfloor)
\label{eq:grid_convert}
\end{equation}
, where we set grid cell size = 0.1.
eCAR calculates viewing grid cells $C^L = \{c_1,c_2,...,c_{Nlc}\}$ in the frame from sampled 3D points.
eCAR traverses grid cells to obtain virtual objects $V^L$. 
eCAR communicates $G^L$ to the device. 

Constructing a local graph, eCAR communicates this graph to the device.
To reduce the local map size, we suggest map point structure decoupling.
This approach synchronizes map point information such as ID and 3D position without descriptor to match with a feature point.
A descriptor of map points in the edge server is calculated from a lossy image.
Also, a descriptor size is 32 bytes.
The total descriptor size of a keyframe that detects 1000 key points is 32KB.
Transmitting this information approximately three times per second from multiple devices strains the network.
According to ORB-SLAM\cite{orb_slam1}, the local mapping module updates the medium descriptor and the normal vector and maximum and minimum distance to increase robustness when generating a new keyframe.
This additional information is also synchronized to devices.

\begin{figure}[t]
\begin{center}
\includegraphics[width=1.0\linewidth]{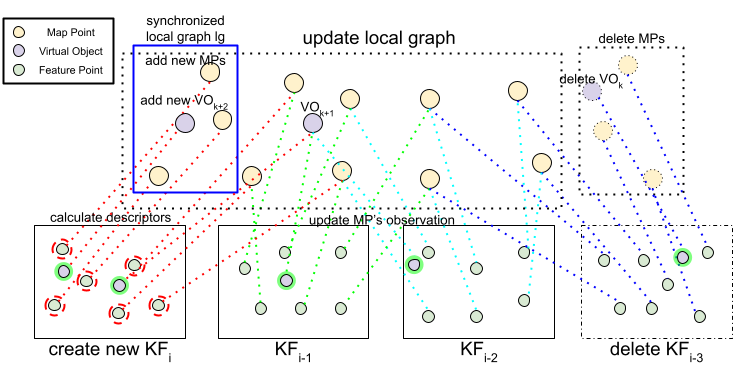}
\end{center}
\caption{Local Graph Update in the device. A device manages a local graph using the queue of keyframes. The device selects the image sent to the server as the key frame for coordinate alignment.}
\label{fig:local_graph_sync}
\end{figure}
In this research, the device analyzes the received local graph to generate the keyframe, considering each frame transmitted to the server as a keyframe. The received local graph includes recently tracked map points in the server. The device uses a queue to maintain several fixed keyframes to manage memory efficiently, and local maps are constructed using keyframes within this queue.
Figure \ref{fig:local_graph_sync} illustrates the local map update process on the device. 
The received local map from the server consists of map point information with separated descriptors corresponding to the images. 
In addition to map point information such as 3D position and ID, the information sent by the server includes 2D positions for calculating new descriptors, keypoint angles, and octave information. 
The device utilizes this information to create keyframes. 
The generated keyframe information maintains a visibility graph for synchronizing the SLAM map state. When a new keyframe $k_{i}$ is generated, corresponding map points and observations are added. Simultaneously, the oldest keyframe $k_{i-3}$ is removed, removing map point observations within that keyframe. 
If the number of observations counts for a map point becomes zero, the device deletes that map point. 
Through this process, the device continuously updates the local map.
In this paper, the device's queue size is 10.

Local graph synchronization operates at the back-end of the device.
Pose tracking operates at the front-end of the device.
These two modules maintain augmented reality consistency by visualizing 3D virtual objects while minimizing the impact of network communication delays.
\subsection{Virtual Object Connection with Grid Cell}
eCAR generates and transmits messages from the user's device to manipulate virtual objects for synchronization among the edge server and devices. 
This message includes the virtual object's ID, 3D position, and operation keyword. 
There are two operation keywords: Registration $I_R(v)$ and Manipulation $I_M(v)$.

The registration process is as follows. 
First, when the device detects a touch input on the screen, it calculates the 3D position based on that location. This calculation considers the plane information at the touched position and the device's orientation, accurately calculating the 3D position using equ (\ref{equ:p_recon}).
While the calculation is ongoing, the device analyzes the points where the ray intersects and selects the keyword to include in the message. 
The Registration keyword is chosen if the intersection point is on a plane. 
If the intersection point corresponds to an existing virtual object, the Manipulation keyword is selected. 
The ID of the intersecting virtual object is also included in the message.
Finally, the generated message is transmitted from the device to the server. 
This process repeats for each frame until the touch input ends, reflecting the user's manipulations on the server.

Upon receiving interaction messages from devices, the server updates the graph. 
First, the server interprets the received message, extracting the virtual object's ID, 3D position, and operation keyword. 
Using the virtual object's ID, the server locates it in the graph and updates its 3D position or creates a new virtual object.

Next, the server connects the virtual object to the grid cell using equ( \ref{eq:grid_convert}). 
Through this connection, the server can include virtual objects in the graph transmitted from the server to the device. 
For example, when the position of a virtual object changes from one end of a corridor to the other, the grid connection allows devices at both ends of the corridor to see the virtual object.
\begin{figure}[!t]
\centering
    \subfloat[Virtual Object Registration\label{fig:registration}]{%
       \includegraphics[width=0.5\linewidth]{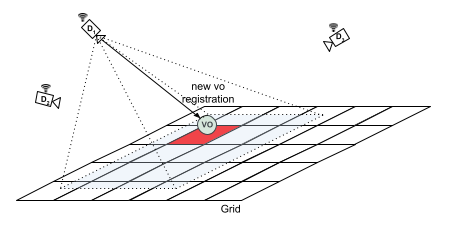}}
    \hfill
    \subfloat[Virtual Object Update\label{fig:manipulation}]{%
       \includegraphics[width=0.5\linewidth]{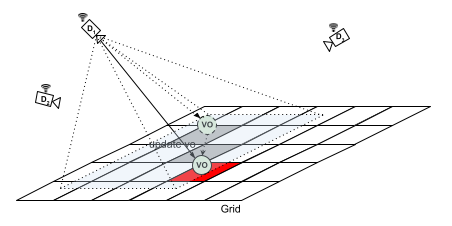}}
\caption{Example of Virtual Object Manipulation.}
\label{fig:grid_vo_connection}
\end{figure}

Figure \ref{fig:grid_vo_connection} describes the connection between a virtual object and a grid cell in the graph of the edge server. 
equation (\ref{eq:back_proj_lay}) calculates a 3D point on the floor with the device's touch input.
The device sends an interaction message to the edge server.
After that, equation (\ref{eq:grid_convert}) selects a grid cell for connecting the graph and a virtual object.
The edge server connects the grid cell and the virtual object in the graph.
The edge server re-calculates and re-connects a grid cell when the user manipulates the virtual object in figure \ref{fig:grid_vo_connection}.
As mentioned, eCAR communicates virtual objects that include local graph $G^L$.
In other words, the edge server asynchronously communicates updated virtual object states to devices to minimize network transmission.
\begin{figure}[t]
\begin{center}
\includegraphics[width=1.0\linewidth]{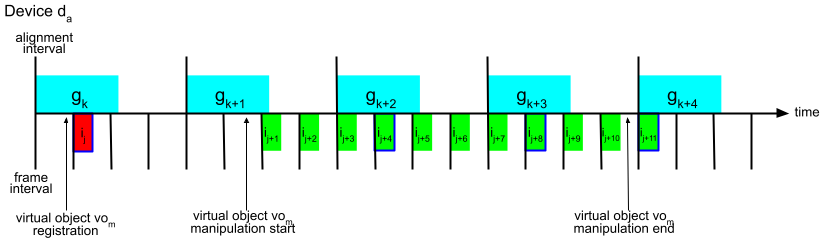}
\end{center}
\caption{Example of Asynchronous Communication. The sky blue square is a graph synchronization latency. The upper column is the data communication interval for matching in the device, and the lower column is the frame interval. In the example, the matching interval is assumed to be four frames. The red rectangle is the virtual object registration interaction latency, and the green rectangle is the virtual object update interaction latency. A blue borderline of the square is a status message visualized on the device $d_a$. eCAR synchronizes virtual objects by each device in graph update.}
\label{fig:async_example}
\end{figure}
Figure \ref{fig:async_example} illustrates an example of virtual object management in the graph. 
In the figure, eCAR registers the virtual object $v_m$ in the graph using the received virtual object interaction message $i_{j}$ on device $d_a$. The graph communication $g_k$ includes map data for tracking and $v_m$ for interaction. Virtual object $v_m$ can be updated anytime by virtual object interaction messages in the graph. Graph communication transmits the most recently updated state of virtual objects to the device.
\section{Experiments}
\subsection{Experimental setup}
\begin{table}[]
\resizebox{\linewidth}{!}{%
{\scriptsize
\begin{tabular}{|l|l|l|}
\hline
Platform & Type      & Hardware Characteristics                                            \\ \hline
Server & desktop & \begin{tabular}[c]{@{}l@{}}AMD Ryzen 9 5950X @ 3.40GHz(16 Cores)\\ 128 GB RAM\end{tabular} \\ \hline
Client   & laptop    & \begin{tabular}[c]{@{}l@{}}AMD Ryzen 9 7940HS @ 4.0GHz(16 Logical Processor)\\ 32 GB RAM\end{tabular}\\ \hline
Client & SM-G980 & \begin{tabular}[c]{@{}l@{}}Qualcomm Snapdragon SM8250\\ 12GB RAM\end{tabular}              \\ \hline
Client   & Pixel6pro & \begin{tabular}[c]{@{}l@{}}Google Tensor\\ 12GB RAM\end{tabular}    \\ \hline
Client   & Pixel7pro & \begin{tabular}[c]{@{}l@{}}Google Tensor G2\\ 12GB RAM\end{tabular} \\ \hline
\end{tabular}%
}
}
\caption{
Hardware setup for eCAR
}
\label{tab:hardware}
\end{table}
Table \ref{tab:hardware} shows hardware lists that are used in this study.
The server sends the identical coordinate system and virtual objects to clients.
Clients receive local graphs and send images for coordinate alignment.
The laptop evaluates the scalability of eCAR with simulation.
To evaluate scalability, we implement a communication module with the same mechanism in a laptop as in mobile devices. 
We operate 20 simulators simultaneously.
We implement the collaborative augmented reality application on a device with Unity Android\cite{unity}.
Also, eCAR communicates data between server and clients with REST API\cite{curl,flask}.

We measure following metrics.
\begin{itemize}
    \item Traffic(byte): the size of the synchronized local graph over the network.
    \item Latency(ms): delay in communication to download the graph after sending the image to the server for graph synchronization.
    \item ATE(cm): Difference between the estimated pose of a frame and the ground truth pose. We evaluate the accuracy of the SLAM map in the edge server and the accuracy of synchronization in the device.
    \item Success Rate(\%): the synchronization success rate of the virtual object from server to device.
\end{itemize}

We communicate with campus wireless.
Since this network is accessible from anywhere in the building, it is suitable for measuring graph synchronization performance due to increased access devices.
Campus wireless network has a download speed of 3.98 MB/s and an upload speed of 3.5 MB/s, as measured by Google.
The upload network bandwidth is 150Mbps, and the download network bandwidth of campus wireless is 600Mbps.
We measured the proposed system performance with campus wireless at weekday work hours.

We evaluate our proposed framework with TUM dataset\cite{tumdataset} and self-collected datasets.
The TUM dataset evaluates the tracking accuracy of the edge server.
Self-produced datasets were recorded with slightly different routes in an indoor corridor environment.
We repeatedly measured each quantitative experiment using these datasets 20 times while increasing the number of devices.
\def\trafficecar{8507.40}
\def\trafficbase{53275.62}
\def\latecarstart{53}
\def\latecarend{69}
\def\latbasestart{63}
\def\latbaseend{111}
\def\serverupload{3}
\def\serverdownload{2}
\def\serverecaralign{17}
\def\serverbasealign{18}
\def\deviceecarupstart{14}
\def\deviceecarupend{26}
\def\deviceecardownstart{10}
\def\deviceecardownend{18}
\def\devicebaseupstart{14}
\def\devicebaseupend{34}
\def\devicebasedownstart{17}
\def\devicebasedownend{47}
\def\voscaleecarmin{56.11}
\def\voscaleecarmax{69.41}
\def\voscalebasemin{28.10}
\def\voscalebasemax{493.85}
\def\longdataset{Corridor dataset}
\def\circledataset{Circle trajectory dataset}
\subsection{Tracking Map synchronization evaluation}
\subsubsection{Local Graph Traffic and Latency Evaluation}
\begin{figure}[!t]
\centering
    \subfloat[Graph Synchronization Latency\label{fig:align_latency}]{%
       \includegraphics[width=0.5\linewidth]{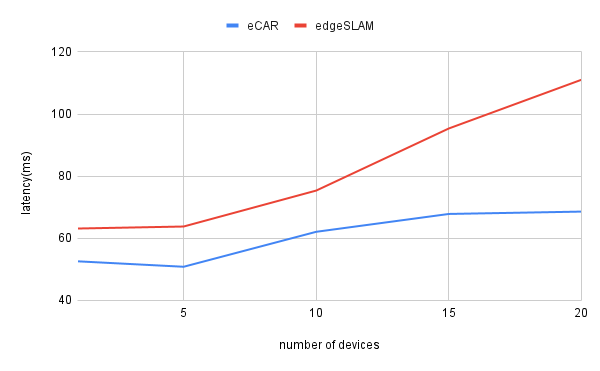}}
   \subfloat[Local Graph Traffic\label{fig:align_traffic}]{%
       \includegraphics[width=0.5\linewidth]{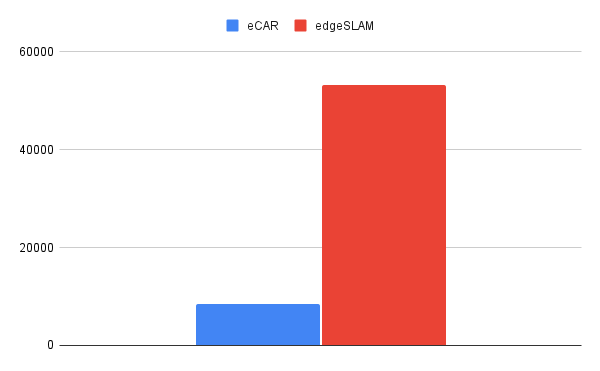}}
\caption{Graph Synchronization Latency and Traffic}
\label{fig:align_eval}
\end{figure}

We measure graph synchronization latency and traffic of eCAR with public and self-corrected datasets.
eCAR sends a local graph, including map data and virtual objects.
The map data in a local graph is reduced by our proposed approaches.
edgeSLAM\cite{edgeSemanticSLAM1} only sends an updated local map after synchronization to reduce network transmission traffic.

Figure \ref{fig:align_traffic} shows the network traffic of eCAR and edgeSLAM.
The average network traffic of eCAR for graph synchronization is \trafficecar bytes.
The average network traffic of edgeSLAM for graph synchronization is \trafficbase bytes.
The map synchronization traffic in eCAR is reduced to about 20\% compared to traffic in the edgeSLAM.

eCAR is lower latency than baseline by small network traffic.
Figure \ref{fig:align_latency} shows measured latency.
The latency of the two methods increases as the number of access devices increases.
eCAR graph synchronization latency is \latecarstart ms when the number of access devices is 1.
This latency increases up to \latecarend when the number of access devices is 20.
The baseline latency varies from \latbasestart ms to \latbaseend ms depending on the number of access devices. 

Collaborative augmented reality applications connect multiple devices to the same network concurrently.
eCAR maintains the local graph status of devices with small-sized map synchronization under bad network performance by increasing devices in the network.
Thus, eCAR's graph synchronization is suitable for communicating data while maintaining network transmission performance in collaborative augmented reality.
\subsubsection{Scalability Evaluation}
In this experiment, we evaluate the scalability of eCAR when connected devices increase.
Co-located devices manipulate and synchronize virtual objects in collaborative augmented reality.
Thus, eCAR's centralized communication for graph synchronization is affected by several connected devices.
To evaluate scalability, we measure processing time in the edge server and devices.
In this experiment, devices record data upload time and download time.
Also, the edge server records image download time and graph upload time and coordinates alignment processing time.

\begin{figure}[!t]
\centering
    \subfloat[Device Processing Time Evaluation\label{fig:align_s_d}]{%
       \includegraphics[width=0.5\linewidth]{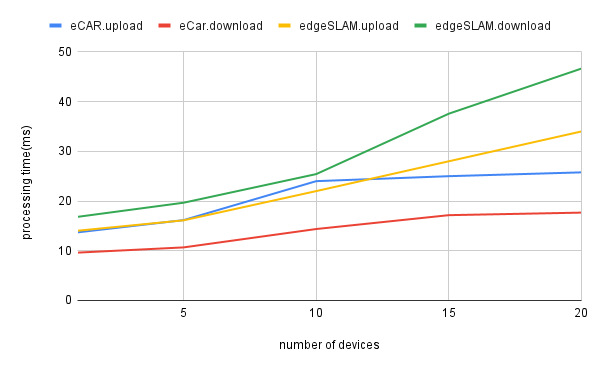}}
   \subfloat[Server Processing Time Evaluation\label{fig:align_s_s}]{%
       \includegraphics[width=0.5\linewidth]{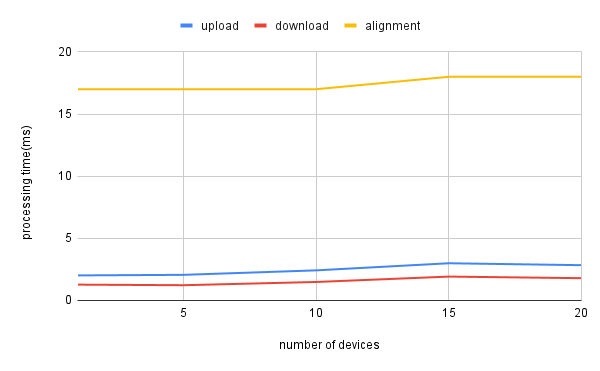}}
\caption{Alignment Scalability Evaluation}
\label{fig:align_scale}
\end{figure}

Figure \ref{fig:align_scale} shows evaluation results.
As shown in this figure, the increase in the network transmission time of eCAR is less than the increase in the network transmission time of the edgeSLAM.
Figure \ref{fig:align_s_d} shows measured upload and download time in devices.

According to previous experimental results, eCAR and edgeSLAM deteriorate network conditions as the number of access devices increases, increasing upload time and download time.
On the other hand, servers are hardly affected by an increase in the number of access devices.
Figure \ref{fig:align_s_s} shows the upload time, download time, and coordinate alignment time measured by the server.
Servers have similar total processing times for both eCAR and edgeSLAM.
In addition, the server's processing time hardly increases up to 20 access devices.
In summary, graph communication latency is proportional to the increase in data communication time of a device.

Analyzing the experimental results, graph synchronization traffic in the edgeSLAM affects latency in collaborative augmented reality, even if an average size is 50000 bytes.
Collaborating devices connect to the same Wi-Fi network to transmit and receive data.
The input data of the edgeSLAM and the eCAR are identical encoded images, but the output data are different.
The output data size of the eCAR is smaller than the output data size of the edgeSLAM.
Therefore, as the number of access devices increases, graph synchronization of the edgeSLAM deteriorates the network condition.
In conclusion, the size of the output data must be small to maintain the coordinate system in collaborative augmented reality, where multiple devices are connected to the same network.
\subsubsection{Tracking Evaluation}
eCAR transmits compressed images to reduce traffic, resulting in lower image quality than the original images. 
Therefore, we evaluate the localization accuracy affected by changes in image quality.
To evaluate localization accuracy, we conduct two experiments. 
First, to evaluate the mapping accuracy of the server, we measure Absolute Trajectory Error (ATE) using a tum dataset. 
Second, to evaluate the graph synchronization of eCAR, we measure ATE between the server and the device using the map received from the server.
We varied the compression quality from 100 to 10 in increments of 10.
Each sequence was repeated 20 times for each compression quality.

\begin{table*}[]
\resizebox{\textwidth}{!}{%
\begin{tabular}{|l|cccccccccccc|}
\hline
 &
  \multicolumn{12}{c|}{Absolute KeyFrame Trajectory Error(cm)} \\ \hline
 &
  \multicolumn{10}{c|}{eCAR} &
  \multicolumn{1}{c|}{edgeSLAM} &
  ORB\_SLAM \\ \hline
\multicolumn{1}{|c|}{JPEG Quality} &
  \multicolumn{1}{c|}{100} &
  \multicolumn{1}{c|}{90} &
  \multicolumn{1}{c|}{80} &
  \multicolumn{1}{c|}{70} &
  \multicolumn{1}{c|}{60} &
  \multicolumn{1}{c|}{50} &
  \multicolumn{1}{c|}{40} &
  \multicolumn{1}{c|}{30} &
  \multicolumn{1}{c|}{20} &
  \multicolumn{1}{c|}{10} &
  \multicolumn{1}{c|}{} &
   \\ \hline
tum1\_floor &
  1.584 &
  2.16 &
  1.826 &
  \textbf{1.471} &
  1.58 &
  1.694 &
  1.841 &
  1.771 &
  1.623 &
  \multicolumn{1}{c|}{2.135} &
  1.523 &
  2.99 \\
tum1\_xyz &
  2.42 &
  0.918 &
  1.103 &
  1.071 &
  1.93 &
  1.854 &
  1.698 &
  1.337 &
  1.402 &
  \multicolumn{1}{c|}{1.285} &
  1.152 &
  \textbf{0.90} \\
tum2\_desk &
  1.373 &
  1.044 &
  1.193 &
  \textbf{0.829} &
  1.32 &
  1.318 &
  1.241 &
  1.481 &
  1.147 &
  \multicolumn{1}{c|}{1.832} &
  0.835 &
  0.88 \\
tum2\_desk\_with\_person &
  0.68 &
  0.8 &
  0.865 &
  0.689 &
  \textbf{0.628} &
  0.69 &
  0.813 &
  0.946 &
  1.027 &
  \multicolumn{1}{c|}{0.989} &
  0.718 &
  0.63 \\
tum2\_xyz &
  0.227 &
  0.24 &
  0.23 &
  \textbf{0.22} &
  0.225 &
  0.214 &
  0.236 &
  0.253 &
  0.256 &
  \multicolumn{1}{c|}{0.258} &
  0.223 &
  0.3 \\
tum3\_long\_office &
  1.651 &
  1.102 &
  1.305 &
  1.216 &
  1.441 &
  1.098 &
  1.254 &
  \textbf{1.041} &
  1.176 &
  \multicolumn{1}{c|}{2.124} &
  1.212 &
  3.45 \\
tum3\_str\_tex\_far &
  1.009 &
  1.0 &
  0.853 &
  0.972 &
  0.968 &
  0.986 &
  0.98 &
  0.981 &
  0.911 &
  \multicolumn{1}{c|}{0.99} &
  0.968 &
  \textbf{0.77} \\
tum3\_str\_tex\_near &
  1.298 &
  1.413 &
  \textbf{1.255} &
  1.609 &
  1.542 &
  1.282 &
  1.264 &
  1.483 &
  1.524 &
  \multicolumn{1}{c|}{2.428} &
  1.614 &
  1.39 \\ \hline
\end{tabular}%
}
\caption{
Absolute Trajectory Error Evaluation(Server).
}
\label{tab:tum_localization2}
\end{table*}
To evaluate the server map creation accuracy concerning compression quality changes, we generated maps from images with varying compression qualities. 
Table \ref{tab:tum_localization2} presents the ATE of maps created from sequences in eCAR's TUM dataset\cite{tumdataset}. 
Upon reviewing the results, it is evident that eCAR's server map, generated from degraded images, effectively maintains localization performance.
Even as compression quality decreases, ATE generally remains consistent. Specifically, the average compression quality with the smallest ATE is around 70. 
Analyzing the experimental results, the loss compression algorithm mainly removes the high-frequency area of the image where the edge area is distributed.
In other words, this processing is similar to applying image blurring, especially at feature points where significant brightness changes are mainly detected.
Therefore, even with reduced image quality, eCAR robustly detects and matches feature points, ensuring effective maintenance of mapping accuracy.

\begin{table}[]
\resizebox{0.8\columnwidth}{!}{%
\begin{tabular}{|ccc|}
\hline
\multicolumn{3}{|c|}{Absolute KeyFrame Trajectory Error(cm)}                          \\ \hline
\multicolumn{1}{|c|}{sequence}                 & \multicolumn{1}{c|}{eCAR} & edgeSLAM \\ \hline
\multicolumn{1}{|c|}{tum1\_floor}              & \multicolumn{1}{c|}{1.17} & 1.18     \\ \hline
\multicolumn{1}{|c|}{tum1\_xyz}                & \multicolumn{1}{c|}{0.76} & 0.78     \\ \hline
\multicolumn{1}{|c|}{tum2\_desk}               & \multicolumn{1}{c|}{0.47} & 0.49     \\ \hline
\multicolumn{1}{|c|}{tum2\_desk\_with\_person} & \multicolumn{1}{c|}{0.10} & 0.11     \\ \hline
\multicolumn{1}{|c|}{tum2\_xyz}                & \multicolumn{1}{c|}{0.51} & 0.50     \\ \hline
\multicolumn{1}{|c|}{tum3\_long\_office}       & \multicolumn{1}{c|}{0.15} & 0.21     \\ \hline
\multicolumn{1}{|c|}{tum3\_str\_tex\_far}      & \multicolumn{1}{c|}{0.12} & 0.15     \\ \hline
\multicolumn{1}{|c|}{tum3\_str\_tex\_near}     & \multicolumn{1}{c|}{0.13} & 0.14     \\ \hline
\end{tabular}%
}
\caption{
Absolute Trajectory Error Evaluation(Device).
}
\label{tab:tum_device_localization}
\end{table}
In this section, we evaluate the performance of map data synchronization in the graph by measuring the discrepancy between the device's estimated position and the position estimated by the server. 
For evaluation purposes, the device reconstructs its pose using the graph information received from the server. 
eCAR transmits map points with separated descriptors that correspond to recently tracked points. 
The device then reconstructs the map using this information. 
edgeSLAM involves transmitting the server's local map information as is. 
Table \ref{tab:tum_device_localization} presents the ATE measurement results between the device and the server.
Both methods achieve location measurements with errors of almost 0.5 cm or less. 
As shown in the table, there is no significant difference in accuracy between eCAR's graph synchronization and edgeSLAM graph synchronization.
In conclusion, eCAR trade-off network traffic and localization accuracy to synchronize the graph more frequently. 
eCAR's graph synchronization is faster in terms of latency due to its smaller network traffic than the baseline. 
With minimal impact on latency as the number of connected devices increases, eCAR remains resilient. 
Moreover, even under poor network conditions, eCAR maintains low latency for graph synchronization. 
Despite these advantages, eCAR demonstrates localization accuracy that is nearly equivalent to traditional local map synchronization. 
Therefore, the proposed map data communication is well-suited for collaborative augmented reality.
\subsection{Virtual object synchronization evaluation}
\subsubsection{Scalability Evaluation}
\def\voscaleears{56.11ms }
\def\voscalebases{28.10ms }
\def\voscaleearten{69.41ms }
\def\voscalebaseten{493.85ms }
We measure virtual object synchronization latency for scalability evaluation.
After matching the coordinate system, the eCAR shares the registered virtual object in the graph synchronization process.
Baseline directly shares virtual objects with adjacent devices.
Virtual object sharing increases the sharing time as the number of access devices increases\cite{car_quad}.
Therefore, we measure the virtual object sharing latency while increasing the number of access devices participating in collaborative augmented reality.

\begin{figure}[t]
\begin{center}
\includegraphics[width=1.0\linewidth]{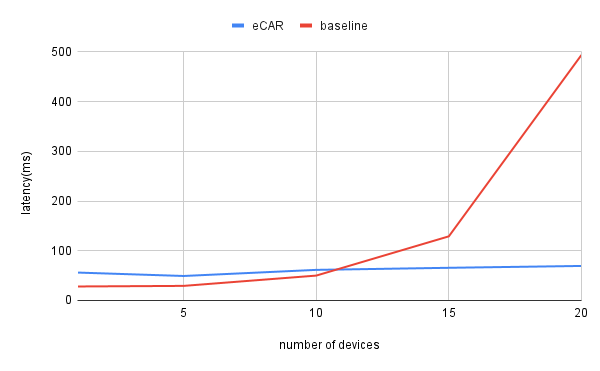}
\end{center}
\caption{Virtual Object Scalability Evaluation}
\label{fig:vo_scale}
\end{figure}
Figure \ref{fig:vo_scale} shows the synchronization performance.
The eCAR's latency is \voscaleears when the access device is 1.
The baseline's latency is \voscalebases when the connection device is 1.
When the access device is less than 10 in the same network, the baseline's virtual object sharing latency is lower than the latency of eCAR.
However, the difference in virtual object latency between the baseline and eCAR decreases as the number of connected devices increases.
When the number of access devices exceeds 15, eCAR clearly synchronizes virtual objects faster than the baseline.
In other words, eCAR's latency is less affected due to an increase in the number of access devices than the baseline.
For a baseline, the latency change is small below a specific number of access devices, but when the number of access devices exceeds 10, the latency increases rapidly.
The reason is that eCAR regulates the number of communications by including virtual object sharing in graph synchronization.
On the other hand, since virtual object direct communication directly shares changes with adjacent devices, the number of communications for synchronization increases as the number of access devices increases.
Nevertheless, when the number of access devices is below a certain level, direct communication has a lower latency than indirect communication.
On the other hand, eCAR reflects the virtual object registered by a device to the edge server and transmits the reflected information together when other devices match the coordinate system.
Therefore, eCAR is unsuitable for applications such as games that send and receive real-time messages. 
Still, eCAR is suitable for tasks such as fine-tuning the location while simultaneously viewing the virtual object from multiple angles in collaboration to draw pictures or match the virtual object to the correct location.
\subsubsection{Synchronization Range Evaluation}
\def\vovisearlong{100\%}
\def\vovisearhalf{100\%}
\def\vovisbaselong{99\%}
\def\vovisbasehalf{99\%}
In this section, we quantitatively evaluate the virtual object synchronization accuracy.
To evaluate this, we measure the synchronization success rate while varying the distance and angle.
eCAR indirectly connects by adding grids between a co-visibility graph and virtual objects for virtual object sharing.
eCAR shares virtual objects connected to the grid seen in the current device frame.
On the other hand, the baseline connects the virtual object directly to the keyframe of the co-visibility graph.
This shares virtual objects connected to the keyframe of the local map constructed in the edge server coordinate system matching.

\begin{figure}[!t]
\centering
    \subfloat[Long corridor scene\label{fig:vo_acc_dist}]{%
       \includegraphics[width=0.5\linewidth]{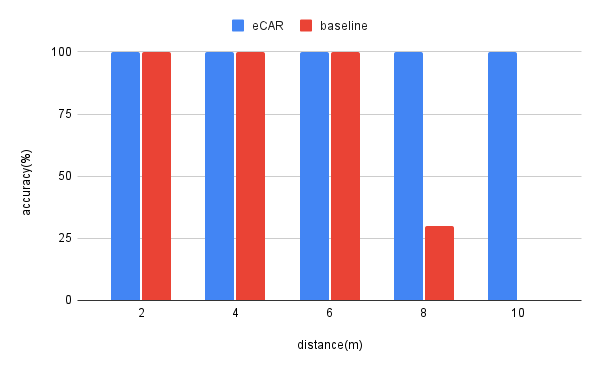}}
   \subfloat[Half circle scene\label{fig:vo_acc_ang}]{%
       \includegraphics[width=0.5\linewidth]{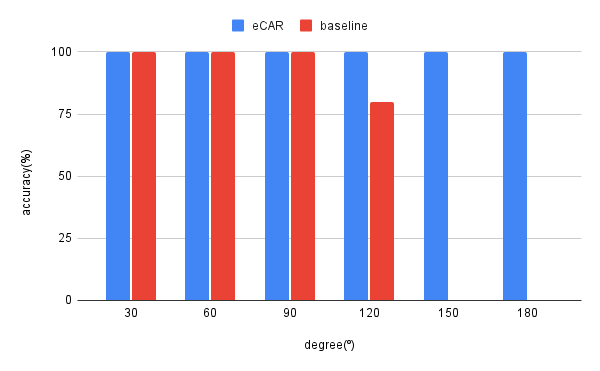}}
\caption{Virtual Object Synchronization Range Evaluation}
\label{fig:vo_acc}
\end{figure}s
We conduct experiments with two self-collect sequences.
The corridor dataset is a sequence that travels about 30m from the end of the corridor to the opposite side. 
We measure the success rate of virtual object sharing while changing the device's position on the corridor.
Figure \ref{fig:vo_acc_dist} shows the success rate with the corridor sequence.
Upon checking the experimental results, eCAR shares virtual objects without being affected by distance.
On the other hand, the baseline rarely shares the virtual object when the distance between the virtual object and the device is approximately 10m or more.
A circle trajectory dataset is a sequence that moves around a specific point 180\degree while looking at a floor.
We calculate the angle between the device that registered the virtual object and the other device that shares the virtual object based on the virtual object.
We measure the limitation range of degrees for sharing virtual objects with this sequence.

Figure \ref{fig:vo_acc_ang} shows a virtual object synchronization success rate with a circle trajectory dataset.
To measure the success rate, the device moves in a half circle around the virtual object.
Upon checking the experimental results, eCAR shares virtual objects without angle limitations.
On the other hand, the baseline dramatically drops in success rate above 90 degrees.
\subsubsection{Qualitative evaluation of virtual object visualization}
We qualitatively evaluate the graph update and synchronization of the edge server when a user manipulates the virtual object.
If a user modifies the position of a virtual object, the edge server updates the graph connection of this virtual object. 
In previous experiments, we quantitatively evaluated the virtual object registration limitations range.
We have confirmed that virtual object sharing using keyframes has physical limitations range.
This experiment evaluates the graph update of the edge server when the user manipulates a virtual object already registered in the edge server.

\begin{figure}[!t]
\centering
    \subfloat[eCAR\label{fig:vo_update_d_g}]{%
       \includegraphics[width=1\linewidth]{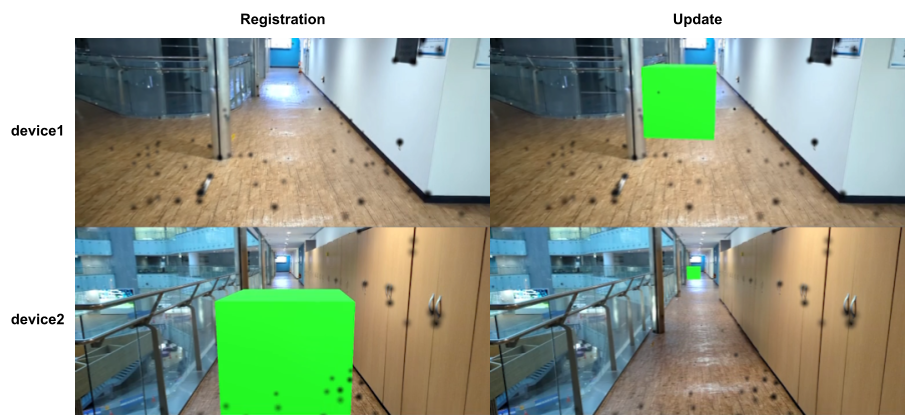}}
   \vfill
   \subfloat[baseline\label{fig:vo_update_d_k}]{%
       \includegraphics[width=1\linewidth]{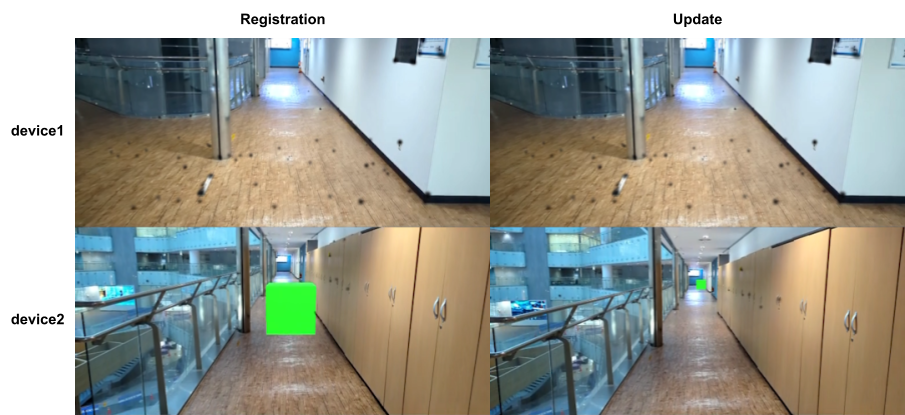}}
\caption{Virtual Object Update Evaluation with long trajectory}
\label{fig:vo_update_long}
\end{figure}

We set device one at the middle of the corridor and device two at the end of the corridor.
First, we register a virtual object in the middle of the corridor.
After that, the registered virtual object is moved to the end of the corridor.
Lastly, we evaluate that the updated virtual object is visualized on device one.
Figure \ref{fig:vo_update_long} shows virtual object update with \longdataset.
Figure \ref{fig:vo_update_d_g} shows eCAR's update results.
Figure \ref{fig:vo_update_d_k} shows baselines' update results.

\begin{figure}[!t]
\centering
    \subfloat[eCAR\label{fig:vo_update_h_g}]{%
       \includegraphics[width=1\linewidth]{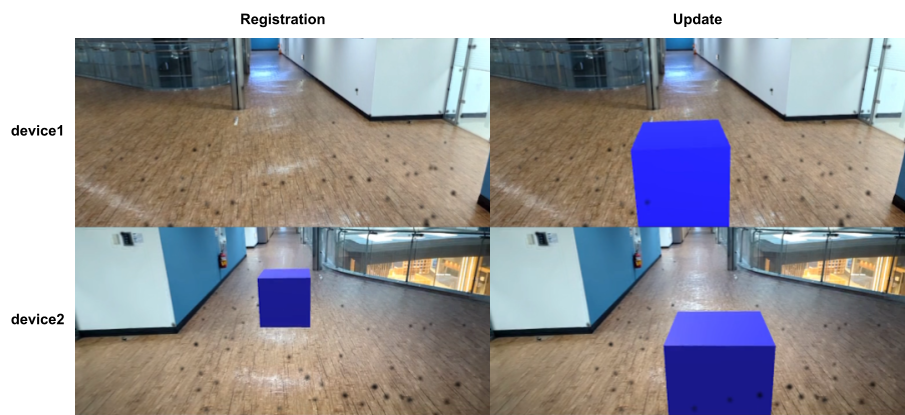}}
    \vfill
    \subfloat[baseline\label{fig:vo_update_h_k}]{%
       \includegraphics[width=1\linewidth]{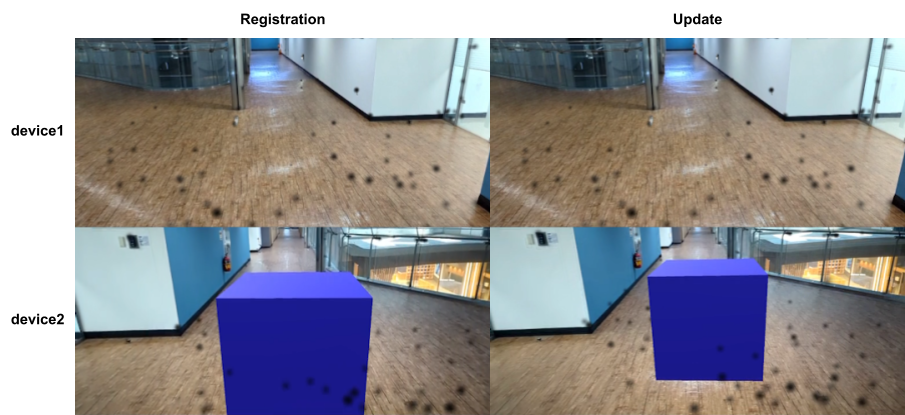}}
\caption{Virtual Object Update Evaluation with half sphere trajectory}
\label{fig:vo_update_half}
\end{figure}

We set up the two devices facing each other in the corridor.
First, the device two registers the virtual object to be visible only on the device 2.
After that, the virtual object is moved in front of the device one.
Lastly, we evaluate that the updated virtual object is visualized on device one.
Figure \ref{fig:vo_update_half} shows virtual object update with \circledataset.
Figure \ref{fig:vo_update_h_g} shows eCAR's update results.
Figure \ref{fig:vo_update_h_k} shows baseline update results.

Analyzing the experimental results, grid-based virtual object connection updates the virtual object without any range limitation.
On the other hand, keyframe-based virtual object connection has physical range limitations in both virtual object registration and manipulation.
In a co-visibility graph, the edge of the graph node between keyframes is linked to a map point observed simultaneously with each other.
If the distance increases during mapping by XYZ motion, there is a high possibility that map points will be shared because they continue to look in the same direction.
On the other hand, the circle sequence moves as the camera rotates along the half-circle.
Frames moving along the half-circle have less area where the view overlaps.
Therefore, the number of shared map points decreases, making it difficult to link keyframe nodes in a co-visibility graph.
Therefore, we propose a simplified representation of the map of a co-visibility graph as a grid to efficiently share the 3D positional change of a virtual object caused by a user's virtual object manipulation in collaboration.
\subsubsection{Collaborative Augmented Reality Application Implementation}
In this section, we qualitatively evaluate graph synchronization among multiple devices in an indoor environment.
For evaluation, we implement a collaborative augmented reality drawing application that paints floors and walls with a virtual brush.
This application represents touch inputs as a virtual line and sends this virtual line for registration to the edge server.
These painted virtual lines are generated until a touch input is finished.
The virtual line has the following information: id, start point, end point, RGB color, line width, and normal vector.
We evaluate that adjacent devices download and visualize these virtual lines without bottleneck.

We record video sequences of four mobile devices in a corridor.
CAR drawing applications are run on four devices with recorded images to experiment with whether virtual paint objects can be shared simultaneously.
We set four devices in the same direction from the corridor.
Devices 1 and 2 are located side by side at the end of the corridor.
Devices 3 and 4 are located side by side in the middle of the corridor.

\begin{figure}
    \centering
    \includegraphics[width=1\linewidth]{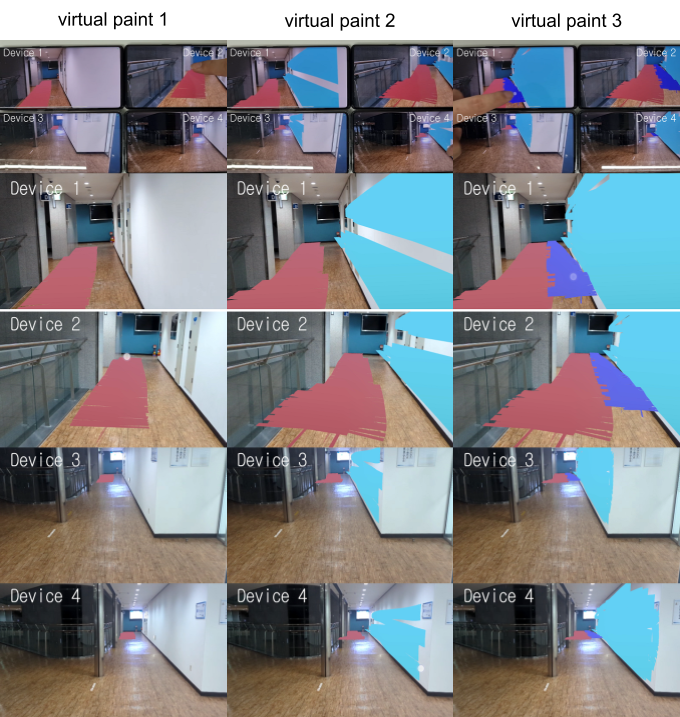}
    \caption{Collaborative Drawing Application}
    \label{fig:draw_app}
\end{figure}
Figure \ref{fig:draw_app} shows the result of drawing virtual paints on the floor and wall.
The columns in the figure show the virtual paint input. 
A white circle in a device is a touch input of a screen at the first row.
The rows in the figure show that the virtual paint is visualized on each instrument simultaneously.
The first column shows the user applying red paint to the floor at the end of the corridor on Device 2.
The second column shows the user painting sky blue paint on the wall at the end of the corridor on Device 4.
The third column shows the user applying blue paint to the floor at the end of the corridor on Device 1. 
As shown in the above figure, virtual paint is visualized on the identical position of the floor or wall at the same line width in each column.
Also, these virtual paints are seamlessly registered on the floor or wall.

Analyzing the experimental results, eCAR quickly and accurately shares augmented reality experiences with adjacent devices in a large indoor environment.
As known in the experiment results, eCAR offers proper local graphs to each device.
Also, the edge server synchronizes the huge virtual objects generated by user input without evident delay.
Through collaborative augmented reality drawing applications, eCAR's virtual object-sharing mechanism enables various collaborative applications, such as navigation, interior design, or message sharing, in large indoor environments such as museums and shopping malls.
\section{Conclusion and Future Works} 
In this work, we proposed a collaborative augmented reality framework, eCAR, that provides the same augmented reality experience between multiple devices.
eCAR continuously synchronizes the latest graph information of the edge server with the device.
Through this synchronization, devices maintain SLAM map points and virtual object states.
Furthermore, eCAR has designed a graph-grid-virtual object structure to share virtual objects and interact with real space.
A device registers the virtual object in the real environment through this structure, and the edge server shares the virtual object with other devices.
Finally, eCAR maintains low graph synchronization latency by minimizing traffic size by proposed approaches that minimize map data.

The limitation of the proposed research is that when the network situation worsens due to the increase of the access device, the end-to-end latency increases in the device.
To address this, we will study hybrid communication that controls data directly between adjacent devices based on eCAR's centralized communication mechanism.
Furthermore, eCAR implements a grid connected to a graph that briefly represents the main planar areas, such as floors and walls in the indoor environment.
Due to the briefly represented grid, eCAR registered and updated virtual objects intuitively in a real-world environment.
However, eCAR is not suitable for representing real objects.
Therefore, we will combine 3D dense object reconstruction and tracking modules to study interaction with real objects.
Finally, we will study connecting the graph structure of eCAR with a scene graph to efficiently represent AR and VR and synchronize the states of real and virtual objects.
In the future, eCAR will provide users with various augmented reality experiences in large indoor environments such as museums and department stores.

\clearpage 
\bibliographystyle{format/vgtc/abbrv-doi}

\begin{thebibliography}{10}

\bibitem{ceres_solver}
S.~Agarwal and K.~Mierle.
\newblock Ceres solver: Tutorial \& reference.
\newblock {\em Google Inc}, 2(72):8, 2012.

\bibitem{uwb}
G.~R. Aiello and G.~D. Rogerson.
\newblock Ultra-wideband wireless systems.
\newblock {\em IEEE microwave magazine}, 4(2):36--47, 2003.

\bibitem{edgeSLAM2}
A.~J.~B. Ali, M.~Kouroshli, S.~Semenova, Z.~S. Hashemifar, S.~Y. Ko, and K.~Dantu.
\newblock Edge-slam: Edge-assisted visual simultaneous localization and mapping.
\newblock {\em ACM Transactions on Embedded Computing Systems}, 22(1):1--31, 2022.

\bibitem{freear}
K.~Apicharttrisorn, J.~Chen, V.~Sekar, A.~Rowe, and S.~V. Krishnamurthy.
\newblock Breaking edge shackles: Infrastructure-free collaborative mobile augmented reality.
\newblock In {\em Proceedings of the 20th ACM Conference on Embedded Networked Sensor Systems}, pp. 1--15, 2022.

\bibitem{ARKit}
Apple.
\newblock Creating a multiuser ar experience, 2018.

\bibitem{edgeSLAM1}
A.~J. Ben~Ali, Z.~S. Hashemifar, and K.~Dantu.
\newblock Edge-slam: edge-assisted visual simultaneous localization and mapping.
\newblock In {\em Proceedings of the 18th International Conference on Mobile Systems, Applications, and Services}, pp. 325--337, 2020.

\bibitem{car1}
A.~Butz, T.~Hollerer, S.~Feiner, B.~MacIntyre, and C.~Beshers.
\newblock Enveloping users and computers in a collaborative 3d augmented reality.
\newblock In {\em Proceedings 2nd IEEE and ACM International Workshop on Augmented Reality (IWAR'99)}, pp. 35--44. IEEE, 1999.

\bibitem{orb_slam3}
C.~Campos, R.~Elvira, J.~J.~G. Rodr{\'\i}guez, J.~M. Montiel, and J.~D. Tard{\'o}s.
\newblock Orb-slam3: An accurate open-source library for visual, visual--inertial, and multimap slam.
\newblock {\em IEEE Transactions on Robotics}, 37(6):1874--1890, 2021.

\bibitem{wifi_direct}
D.~Camps-Mur, A.~Garcia-Saavedra, and P.~Serrano.
\newblock Device-to-device communications with wi-fi direct: overview and experimentation.
\newblock {\em IEEE wireless communications}, 20(3):96--104, 2013.

\bibitem{edgeSemanticSLAM2}
H.~Cao, J.~Xu, D.~Li, L.~Shangguan, Y.~Liu, and Z.~Yang.
\newblock Edge assisted mobile semantic visual slam.
\newblock {\em IEEE Transactions on Mobile Computing}, 2022.

\bibitem{marvel}
K.~Chen, T.~Li, H.-S. Kim, D.~E. Culler, and R.~H. Katz.
\newblock Marvel: Enabling mobile augmented reality with low energy and low latency.
\newblock In {\em Proceedings of the 16th ACM Conference on Embedded Networked Sensor Systems}, pp. 292--304, 2018.

\bibitem{carkeyfactor3}
Z.~Chen, W.~Hu, J.~Wang, S.~Zhao, B.~Amos, G.~Wu, K.~Ha, K.~Elgazzar, P.~Pillai, R.~Klatzky, et~al.
\newblock An empirical study of latency in an emerging class of edge computing applications for wearable cognitive assistance.
\newblock In {\em Proceedings of the Second ACM/IEEE Symposium on Edge Computing}, pp. 1--14, 2017.

\bibitem{edgear1}
Y.~Cheng.
\newblock Edge caching and computing in 5g for mobile augmented reality and haptic internet.
\newblock {\em Computer Communications}, 158:24--31, 2020.

\bibitem{bundlefusion}
A.~Dai, M.~Nie{\ss}ner, M.~Zollh{\"o}fer, S.~Izadi, and C.~Theobalt.
\newblock Bundlefusion: Real-time globally consistent 3d reconstruction using on-the-fly surface reintegration.
\newblock {\em ACM Transactions on Graphics (ToG)}, 36(4):1, 2017.

\bibitem{cg_vis_graph}
M.~De~Berg.
\newblock Computational geometry: algorithms and applications.
\newblock chap.~15, pp. 323--331. Springer Science \& Business Media, 2000.

\bibitem{slam_share}
A.~Dhakal, X.~Ran, Y.~Wang, J.~Chen, and K.~Ramakrishnan.
\newblock Slam-share: visual simultaneous localization and mapping for real-time multi-user augmented reality.
\newblock In {\em Proceedings of the 18th International Conference on emerging Networking EXperiments and Technologies}, pp. 293--306, 2022.

\bibitem{ear}
N.~Didar and M.~Brocanelli.
\newblock ear: an edge-assisted and energy-efficient mobile augmented reality framework.
\newblock {\em IEEE Transactions on Mobile Computing}, 2022.

\bibitem{dso}
J.~Engel, V.~Koltun, and D.~Cremers.
\newblock Direct sparse odometry.
\newblock {\em IEEE transactions on pattern analysis and machine intelligence}, 40(3):611--625, 2017.

\bibitem{lsdslam}
J.~Engel, T.~Sch{\"o}ps, and D.~Cremers.
\newblock Lsd-slam: Large-scale direct monocular slam.
\newblock In {\em Computer Vision--ECCV 2014: 13th European Conference, Zurich, Switzerland, September 6-12, 2014, Proceedings, Part II 13}, pp. 834--849. Springer, 2014.

\bibitem{edge_bottleneck}
M.~Erol-Kantarci and S.~Sukhmani.
\newblock Caching and computing at the edge for mobile augmented reality and virtual reality (ar/vr) in 5g.
\newblock In {\em Ad Hoc Networks: 9th International Conference, AdHocNets 2017, Niagara Falls, ON, Canada, September 28--29, 2017, Proceedings}, pp. 169--177. Springer, 2018.

\bibitem{ransac}
M.~A. Fischler and R.~C. Bolles.
\newblock Random sample consensus: a paradigm for model fitting with applications to image analysis and automated cartography.
\newblock {\em Communications of the ACM}, 24(6):381--395, 1981.

\bibitem{svo}
C.~Forster, M.~Pizzoli, and D.~Scaramuzza.
\newblock Svo: Fast semi-direct monocular visual odometry.
\newblock In {\em 2014 IEEE international conference on robotics and automation (ICRA)}, pp. 15--22. IEEE, 2014.

\bibitem{ARCore}
Google.
\newblock Working with anchors, 2018.

\bibitem{cloudanchor}
Google.
\newblock Cloud anchors allow different users to share ar experiences, 2021.

\bibitem{flask}
M.~Grinberg.
\newblock {\em Flask web development: developing web applications with python}.
\newblock " O'Reilly Media, Inc.", 2018.

\bibitem{unity}
J.~K. Haas.
\newblock A history of the unity game engine.
\newblock 2014.

\bibitem{comic}
B.~Han, P.~Pathak, S.~Chen, and L.-F.~C. Yu.
\newblock Comic: A collaborative mobile immersive computing infrastructure for conducting multi-user xr research.
\newblock {\em IEEE Network}, 2022.

\bibitem{car2}
A.~Henrysson, M.~Billinghurst, and M.~Ollila.
\newblock Face to face collaborative ar on mobile phones.
\newblock In {\em Fourth ieee and acm international symposium on mixed and augmented reality (ismar'05)}, pp. 80--89. IEEE, 2005.

\bibitem{curl}
M.~Hostetter, D.~A. Kranz, C.~Seed, C.~Terman, and S.~Ward.
\newblock Curl: a gentle slope language for the web.
\newblock {\em World wide web journal}, 2(2):121--134, 1997.

\bibitem{edgerobotics}
P.~Huang, L.~Zeng, X.~Chen, K.~Luo, Z.~Zhou, and S.~Yu.
\newblock Edge robotics: Edge-computing-accelerated multirobot simultaneous localization and mapping.
\newblock {\em IEEE Internet of Things Journal}, 9(15):14087--14102, 2022.

\bibitem{synchronizar}
K.~Huo, T.~Wang, L.~Paredes, A.~M. Villanueva, Y.~Cao, and K.~Ramani.
\newblock Synchronizar: Instant synchronization for spontaneous and spatial collaborations in augmented reality.
\newblock In {\em Proceedings of the 31st Annual ACM Symposium on User Interface Software and Technology}, pp. 19--30, 2018.

\bibitem{overlay}
P.~Jain, J.~Manweiler, and R.~Roy~Choudhury.
\newblock Overlay: Practical mobile augmented reality.
\newblock In {\em Proceedings of the 13th Annual International Conference on Mobile Systems, Applications, and Services}, pp. 331--344, 2015.

\bibitem{visualprint}
P.~Jain, J.~Manweiler, and R.~Roy~Choudhury.
\newblock Low bandwidth offload for mobile ar.
\newblock In {\em Proceedings of the 12th International on Conference on emerging Networking EXperiments and Technologies}, pp. 237--251, 2016.

\bibitem{car3}
H.~Kato, M.~Billinghurst, I.~Poupyrev, K.~Imamoto, and K.~Tachibana.
\newblock Virtual object manipulation on a table-top ar environment.
\newblock In {\em Proceedings IEEE and ACM International Symposium on Augmented Reality (ISAR 2000)}, pp. 111--119. Ieee, 2000.

\bibitem{ptam}
G.~Klein and D.~Murray.
\newblock Parallel tracking and mapping for small ar workspaces.
\newblock In {\em 2007 6th IEEE and ACM international symposium on mixed and augmented reality}, pp. 225--234. IEEE, 2007.

\bibitem{g2o}
R.~K{\"u}mmerle, G.~Grisetti, H.~Strasdat, K.~Konolige, and W.~Burgard.
\newblock g2o: A general framework for graph optimization.
\newblock In {\em 2011 IEEE International Conference on Robotics and Automation}, pp. 3607--3613. IEEE, 2011.

\bibitem{edgeObjectAR}
L.~Liu, H.~Li, and M.~Gruteser.
\newblock Edge assisted real-time object detection for mobile augmented reality.
\newblock In {\em The 25th annual international conference on mobile computing and networking}, pp. 1--16, 2019.

\bibitem{edgear4}
Q.~Liu and T.~Han.
\newblock Dare: Dynamic adaptive mobile augmented reality with edge computing.
\newblock In {\em 2018 IEEE 26th International Conference on Network Protocols (ICNP)}, pp. 1--11. IEEE, 2018.

\bibitem{MetaAnchor}
Meta.
\newblock Spatial anchors overview, 2021.

\bibitem{SpatialAnchor}
Microsoft.
\newblock Azure spatial anchors documentation, 2019.

\bibitem{AzureWayFinding}
Microsoft.
\newblock Anchor relationships and way-finding in azure spatial anchors, 2022.

\bibitem{orb_slam1}
R.~Mur-Artal, J.~M.~M. Montiel, and J.~D. Tardos.
\newblock Orb-slam: a versatile and accurate monocular slam system.
\newblock {\em IEEE transactions on robotics}, 31(5):1147--1163, 2015.

\bibitem{orb_slam2}
R.~Mur-Artal and J.~D. Tard{\'o}s.
\newblock Orb-slam2: An open-source slam system for monocular, stereo, and rgb-d cameras.
\newblock {\em IEEE transactions on robotics}, 33(5):1255--1262, 2017.

\bibitem{kinectfusion}
R.~A. Newcombe, S.~Izadi, O.~Hilliges, D.~Molyneaux, D.~Kim, A.~J. Davison, P.~Kohi, J.~Shotton, S.~Hodges, and A.~Fitzgibbon.
\newblock Kinectfusion: Real-time dense surface mapping and tracking.
\newblock In {\em 2011 10th IEEE international symposium on mixed and augmented reality}, pp. 127--136. Ieee, 2011.

\bibitem{dtam}
R.~A. Newcombe, S.~J. Lovegrove, and A.~J. Davison.
\newblock Dtam: Dense tracking and mapping in real-time.
\newblock In {\em 2011 international conference on computer vision}, pp. 2320--2327. IEEE, 2011.

\bibitem{car4}
T.~Ohshima.
\newblock Ar hockey: Acase study of collaborative augmented reality.
\newblock In {\em Proc. Int. Conf. PatternRecognition}, vol.~2, pp. 1226--1229, 1998.

\bibitem{carkeyfactor2}
K.~S. Park and R.~V. Kenyon.
\newblock Effects of network characteristics on human performance in a collaborative virtual environment.
\newblock In {\em Proceedings IEEE Virtual Reality (Cat. No. 99CB36316)}, pp. 104--111. IEEE, 1999.

\bibitem{avr}
H.~Qiu, F.~Ahmad, F.~Bai, M.~Gruteser, and R.~Govindan.
\newblock Avr: Augmented vehicular reality.
\newblock In {\em Proceedings of the 16th Annual International Conference on Mobile Systems, Applications, and Services}, pp. 81--95, 2018.

\bibitem{sharear}
X.~Ran, C.~Slocum, M.~Gorlatova, and J.~Chen.
\newblock Sharear: Communication-efficient multi-user mobile augmented reality.
\newblock In {\em Proceedings of the 18th ACM Workshop on Hot Topics in Networks}, pp. 109--116, 2019.

\bibitem{spar}
X.~Ran, C.~Slocum, Y.-Z. Tsai, K.~Apicharttrisorn, M.~Gorlatova, and J.~Chen.
\newblock Multi-user augmented reality with communication efficient and spatially consistent virtual objects.
\newblock In {\em Proceedings of the 16th International Conference on emerging Networking EXperiments and Technologies}, pp. 386--398, 2020.

\bibitem{car5}
H.~T. Regenbrecht, M.~Wagner, and G.~Baratoff.
\newblock Magicmeeting: A collaborative tangible augmented reality system.
\newblock {\em Virtual Reality}, 6(3):151--166, 2002.

\bibitem{edgearx5}
P.~Ren, X.~Qiao, Y.~Huang, L.~Liu, C.~Pu, S.~Dustdar, and J.~Chen.
\newblock Edge ar x5: An edge-assisted multi-user collaborative framework for mobile web augmented reality in 5g and beyond.
\newblock {\em IEEE Transactions on Cloud Computing}, 10(4):2521--2537, 2020.

\bibitem{c2tam}
L.~Riazuelo, J.~Civera, and J.~M. Montiel.
\newblock C2tam: A cloud framework for cooperative tracking and mapping.
\newblock {\em Robotics and Autonomous Systems}, 62(4):401--413, 2014.

\bibitem{car_quad}
K.~Ruth, T.~Kohno, and F.~Roesner.
\newblock Secure $\{$Multi-User$\}$ content sharing for augmented reality applications.
\newblock In {\em 28th USENIX Security Symposium (USENIX Security 19)}, pp. 141--158, 2019.

\bibitem{bluetooth}
K.~Sairam, N.~Gunasekaran, and S.~R. Redd.
\newblock Bluetooth in wireless communication.
\newblock {\em IEEE Communications Magazine}, 40(6):90--96, 2002.

\bibitem{multi_uav}
P.~Schmuck and M.~Chli.
\newblock Multi-uav collaborative monocular slam.
\newblock In {\em 2017 IEEE International Conference on Robotics and Automation (ICRA)}, pp. 3863--3870. IEEE, 2017.

\bibitem{ccm_slam}
P.~Schmuck and M.~Chli.
\newblock Ccm-slam: Robust and efficient centralized collaborative monocular simultaneous localization and mapping for robotic teams.
\newblock {\em Journal of Field Robotics}, 36(4):763--781, 2019.

\bibitem{edgear2}
Y.-J. Seo, J.~Lee, J.~Hwang, D.~Niyato, H.-S. Park, and J.~K. Choi.
\newblock A novel joint mobile cache and power management scheme for energy-efficient mobile augmented reality service in mobile edge computing.
\newblock {\em IEEE Wireless Communications Letters}, 10(5):1061--1065, 2021.

\bibitem{tumdataset}
J.~Sturm, N.~Engelhard, F.~Endres, W.~Burgard, and D.~Cremers.
\newblock A benchmark for the evaluation of rgb-d slam systems.
\newblock In {\em Proc. of the International Conference on Intelligent Robot Systems (IROS)}, Oct. 2012.

\bibitem{car6}
Z.~Szalav{\'a}ri, D.~Schmalstieg, A.~Fuhrmann, and M.~Gervautz.
\newblock “studierstube”: An environment for collaboration in augmented reality.
\newblock {\em Virtual Reality}, 3(1):37--48, 1998.

\bibitem{bundle_adjustment}
B.~Triggs, P.~F. McLauchlan, R.~I. Hartley, and A.~W. Fitzgibbon.
\newblock Bundle adjustment—a modern synthesis.
\newblock In {\em Vision Algorithms: Theory and Practice: International Workshop on Vision Algorithms Corfu, Greece, September 21--22, 1999 Proceedings}, pp. 298--372. Springer, 2000.

\bibitem{depthfrommotion}
J.~Valentin, A.~Kowdle, J.~T. Barron, N.~Wadhwa, M.~Dzitsiuk, M.~Schoenberg, V.~Verma, A.~Csaszar, E.~Turner, I.~Dryanovski, et~al.
\newblock Depth from motion for smartphone ar.
\newblock {\em ACM Transactions on Graphics (ToG)}, 37(6):1--19, 2018.

\bibitem{edgeSemanticSLAM1}
J.~Xu, H.~Cao, D.~Li, K.~Huang, C.~Qian, L.~Shangguan, and Z.~Yang.
\newblock Edge assisted mobile semantic visual slam.
\newblock In {\em IEEE INFOCOM 2020-IEEE Conference on Computer Communications}, pp. 1828--1837. IEEE, 2020.

\bibitem{edgear3}
A.~Younis, B.~Qiu, and D.~Pompili.
\newblock Latency-aware hybrid edge cloud framework for mobile augmented reality applications.
\newblock In {\em 2020 17th Annual IEEE International Conference on Sensing, Communication, and Networking (SECON)}, pp. 1--9. IEEE, 2020.

\bibitem{jaguar}
W.~Zhang, B.~Han, and P.~Hui.
\newblock Jaguar: Low latency mobile augmented reality with flexible tracking.
\newblock In {\em Proceedings of the 26th ACM international conference on Multimedia}, pp. 355--363, 2018.

\bibitem{sear}
W.~Zhang, B.~Han, and P.~Hui.
\newblock Sear: Scaling experiences in multi-user augmented reality.
\newblock {\em IEEE Transactions on Visualization and Computer Graphics}, 28(5):1982--1992, 2022.

\bibitem{cars}
W.~Zhang, B.~Han, P.~Hui, V.~Gopalakrishnan, E.~Zavesky, and F.~Qian.
\newblock Cars: Collaborative augmented reality for socialization.
\newblock In {\em Proceedings of the 19th International Workshop on Mobile Computing Systems \& Applications}, pp. 25--30, 2018.

\bibitem{carkeyfactor1}
F.~Zheng.
\newblock {\em Spatio-temporal registration in augmented reality}.
\newblock PhD thesis, The University of North Carolina at Chapel Hill, 2015.

\bibitem{mit_seg}
B.~Zhou, H.~Zhao, X.~Puig, T.~Xiao, S.~Fidler, A.~Barriuso, and A.~Torralba.
\newblock Semantic understanding of scenes through the ade20k dataset.
\newblock {\em International Journal on Computer Vision}, 2018.

\end{thebibliography}

\end{document}